\documentclass{article}

\usepackage{microtype}

\usepackage[utf8]{inputenc}
\usepackage[T1]{fontenc}
\usepackage{hyperref}
\usepackage{url}
\usepackage{booktabs}
\usepackage{amsfonts}
\usepackage{nicefrac}
\usepackage{microtype}
\usepackage{amsmath}
\usepackage{amssymb}
\usepackage{enumitem}
\usepackage{titlesec}
\usepackage{caption}
\usepackage{subcaption}
\usepackage[capitalise]{cleveref}
\usepackage{graphicx}
\usepackage{tikz}
\usepackage{textcomp}
\usepackage[group-separator={,}]{siunitx}
\usepackage{hyperref}

\usepackage[accepted]{icml2021}

\makeatletter
\renewcommand{\Notice@String}{}
\makeatother

\icmltitlerunning{Smaller World Models for Reinforcement Learning}

\DeclareMathOperator{\Cat}{Cat}
\DeclareMathOperator{\Softmax}{Softmax}

\begin{document}

\twocolumn[
\icmltitle{Smaller World Models for Reinforcement Learning}



\icmlsetsymbol{equal}{*}

\begin{icmlauthorlist}
\icmlauthor{Jan Robine}{hhu}
\icmlauthor{Tobias Uelwer}{hhu}
\icmlauthor{Stefan Harmeling}{hhu}
\end{icmlauthorlist}

\icmlaffiliation{hhu}{Department of Computer Science, Heinrich Heine University Düsseldorf, Düsseldorf, Germany}

\icmlcorrespondingauthor{Jan Robine}{jan.robine@hhu.de}

\icmlkeywords{Machine Learning, ICML}

\vskip 0.3in
]



\printAffiliationsAndNotice{}  

\begin{abstract}
 Sample efficiency remains a fundamental issue of reinforcement learning.
Model-based algorithms try to make better use of data by simulating the
environment with a model. We propose a new neural network architecture for
world models based on a vector quantized-variational autoencoder (VQ-VAE)
to encode observations and a convolutional LSTM to predict the next embedding
indices. A model-free PPO agent is trained purely on simulated experience from
the world model. We adopt the setup introduced by \citet{simple}, which only
allows $100K$ interactions with the real environment. We apply our method on
$36$ Atari environments and show that we reach comparable performance to their
\mbox{SimPLe} algorithm, while our model is significantly smaller.
\end{abstract}

\section{Introduction}

Reinforcement learning is a generally applicable framework for finding actions
in an environment that maximizes the sum of rewards. From a probabilistic
perspective, the following probabilities lay the foundations of all
reinforcement learning problems:

The \textit{dynamics} ${p(r_t,s_{t+1} \,|\, s_t,a_t)}$, i.e., the conditional
joint probability of the next reward and state given the current state and
action. This probability defines the environment and is usually unknown. In most
real-world applications the underlying states $s_t$ are not observable, but
instead the environment produces observations $x_t$, which might not contain all
state information. In this case we can only observe
${p(r_t,x_{t+1} \,|\, x_t,a_t)}$.

The \textit{policy} ${p_\theta(a_t | x_t)}$, i.e., the conditional probability
of choosing an action given the current observation, \mbox{where $\theta$} denotes the
parameters of some model, indicating that this probability is not provided by
the environment, but learned. Reinforcement learning methods provide means to
optimize the policy in the sense that the actions that maximize the future sum
of rewards have the highest probability.

\textit{Model-free} algorithms try to optimize the policy
${p_\theta(a_t | x_t)}$ based on real experience from the environment, without
any knowledge of the underlying dynamics. They have shown great success in a
wide range of environments, but usually require a large amount of training data,
i.e., many interactions with the environment. This low sample efficiency makes
them inappropriate for real-world applications in which data collection is
expensive.

\textit{Model-based} algorithms approximate the dynamics
${p_\phi(r_t,x_{t+1} \,|\, x_t,a_t) \approx p(r_t,x_{t+1} \,|\, x_t,a_t)}$,
where $\phi$ denotes some learned parameters, thus building a model of the
environment, which we will call \textit{world model} to differentiate it from
other models. The process of improving the policy using the world model is
called \textit{planning}, but there are two types: first, we can generate
training data by sampling from ${p_\phi(r_t,x_{t+1} \,|\, x_t,a_t)}$ and apply a
model-free algorithm to this \textit{simulated experience}. Second, we can try
to \textit{look ahead} into the future using the world model, starting from the
current observation, in order to dynamically improve the action selection during
runtime.

\paragraph{Contributions.}
In this work we follow a model-based approach and consider the first type of
planning. The ability to generate new experience without acting in the real
environment does effectively increase the sample efficiency. Our main
contributions and insights can be summarized as follows:
\begin{itemize}
  \item VQ-VAE based world models require fewer parameters than previous
    approaches (see \cref{tab:parameters} and \cref{tab:parameters-detail}).
  \item Learning a latent space world model and training an agent on it is
    possible with only 100K interactions (see \cref{tab:results-mean}).
  \item A two-dimensional discrete latent representation combined with a
    dynamics network built from convolutional LSTMs (see
    \cref{fig:architecture}) is sufficient for model-based training in Atari
    games.
\end{itemize}

\clearpage

\section{Related Work}

\paragraph{World Models \citep{world-models}.}
Modeling environments with complex visual observations is a hard task, but
luckily predicting observations on pixel level is not necessary. The authors
introduce latent variables by encoding the high-dimensional observations $x_t$
into lower-dimensional, latent representations $z_t$. For this purpose they use
a variational autoencoder \citep{vae}, which they call the ``vision''
component, that extracts information from the observation at the current time
step. They use an LSTM combined with a mixture density network \citep{mdn} to
predict the next latent variables stochastically. They call this component the
``memory'', that can accumulate information over multiple time steps.

They also condition the policy on the latent variable, which enables them to
stay in latent space, so that decoding back into pixels is not required (except
for learning the representations). This makes simulation more efficient and can
reduce the effect of accumulating errors. In \cref{sec:latent-variables} we
describe in more detail how to integrate latent variables into the dynamics.

They successfully evaluate their architectures on two environments, but it
involves some manual fine-tuning of the policy. They use an evolution strategy
to optimize the policy, which is not suitable for bigger networks. They also use
a non-iterative training procedure, i.e., they randomly collect real experience
only once and then train the world model and the policy. This implies that the
improved policies cannot be used to obtain new experience, and a random policy
has to ensure sufficient exploration, which makes the approach inappropriate for
more complex environments.

\paragraph{Simulated Policy Learning \citep{simple}.}
The authors introduce a new model-based algorithm (SimPLe) and successfully
apply it to Atari games. They use an iterative training procedure, that
alternates between collecting real experience, training the world model, and
improving the policy using the world model.

Another novelty is that they restrict the algorithm to about $100K$
interactions with the real environment, which is considerably less than the
usual $10M$ to $50M$ interactions.

They train the policy using the model-free PPO algorithm \citep{ppo} instead of
an evolution strategy. They use a video prediction model similar to SV2P
\citep{sv2p} and incorporate the input action in the decoding process to predict
the next frame. The latent variable is discretized into a bit vector, that is
predicted autoregressively using an LSTM during inference time. The policy gets
frames as input, which means that decoding the latent variables back into
pixel-level is required.

\newpage
They get very good results on a lot of Atari environments, considering the low
number of interactions.

\paragraph{Dreamer \citep{dreamer} and DreamerV2 \citep{dreamerv2}.}
The architecture of DreamerV2 is closely related to ours. The authors train a
world model with latent representations, such that the agent can operate
directly on the latent variables. One of the main improvements of DreamerV2 over
the model of Dreamer is the discretization of the latent space, as it uses
categorical instead of Gaussian latent variables.

They show that their agent beats model-free algorithms in many Atari games after
$50M$ interactions. This is a quite different goal from our work, where we
attempt to learn as much as possible from only $100K$ interactions. Furthermore,
we base our discrete latent world model on a VQ-VAE and thus discretize the
latent variables using vector quantization.

\section{Discrete Latent Space World Models}

The goal of this work is to extend the idea of \citet{world-models} by using
more sophisticated neural network architectures and evaluating them on Atari
environments with the model-free PPO algorithm instead of evolution strategies.
In particular, we discretize the latent space, but have a fundamentally
different architecture compared to DreamerV2 \citep{dreamerv2}, since our latent
space is two-dimensional. Moreover, we adopt the limitation to $100K$
interactions, the iterative training scheme and some other crucial ideas from
\citet{simple}, which will be explained in later sections.

\subsection{Latent Variables} \label{sec:latent-variables}

\begin{figure}[t]
  \vskip 0.05in
	\centering
	\resizebox{\columnwidth}{!}{
		\begin{tikzpicture}[yscale=0.7]
		\tikzstyle{vertex}=[draw,circle,inner sep=0pt,minimum size=0.8cm]
		\node[vertex] (x) at (0,1.5) {$x_t$};
		\node[vertex] (x2) at (4,1.5) {$x_{t+1}$};
		\node[vertex] (x3) at (8,1.5) {$x_{t+2}$};

		\node[vertex] (z) at (0,0) {$z_t$};
		\node[vertex] (z2) at (4,0) {$z_{t+1}$};
		\node[vertex] (z3) at (8,0) {$z_{t+2}$};

		\node[vertex] (r) at (2.5,0) {$r_t$};
		\node[vertex] (r2) at (6.5,0) {$r_{t+1}$};

		\node[vertex] (y) at (2.5,-1.5) {$y_t$};
		\node[vertex] (y2) at (6.5,-1.5) {$y_{t+1}$};

		\node[vertex] (a) at (0.75,-1.5) {$a_t$};
		\node[vertex] (a2) at (4.75,-1.5) {$a_{t+1}$};

		\node[vertex] (h) at (0,-3) {$h_t$};
		\node[vertex] (h2) at (4,-3) {$h_{t+1}$};
		\node[vertex] (h3) at (8,-3) {$h_{t+2}$};

		\draw[->,dotted] (-0.55,-0.55) -- (z);
		\draw[->,dotted] (-0.75,-3) -- (h);
		\node at (-1.4,-1.5) {$\scriptstyle\cdots$};

		\draw[->] (h) -- (h2);
		\draw[->] (h2) -- (h3);

		\draw[->] (x) -- (z);
		\draw[->] (x2) -- (z2);
		\draw[->] (x3) -- (z3);

		\draw[->] (z) -- (h);
		\draw[->] (z2) -- (h2);
		\draw[->] (z3) -- (h3);

		\draw[->] (z) -- (y);
		\draw[->] (z2) -- (y2);

		\draw[->] (z) -- (a);
		\draw[->] (z2) -- (a2);

		\draw[->] (a) -- (h);
		\draw[->] (a2) -- (h2);

		\draw[->] (a) -- (y);
		\draw[->] (a2) -- (y2);

		\draw[->] (h) -- (y);
		\draw[->] (h2) -- (y2);

		\draw[->] (y) -- (r);
		\draw[->] (y2) -- (r2);

		\draw[->] (y) -- (z2);
		\draw[->] (y2) -- (z3);

		\draw[->,dotted] (h3) -- (8.75,-3);
		\draw[->,dotted] (z3) -- (8.375,-0.75);
		\draw[->,dotted] (8.375,-2.25) -- (h3);

		\node at (8.75,-1.5) {$\scriptstyle\cdots$};
		\end{tikzpicture}
	}
  \vskip 0.1in
	\caption{Graphical model of the world model and the policy, which arises from
		inserting the latent variables and from our independence assumptions.}
	\label{fig:graphical-model}
  \vskip -0.1in
\end{figure}

Similar to \citet{world-models} we approximate the dynamics with the help of
latent variables. The sum rule of probability allows us to introduce a latent
variable $z_t$,
\begin{equation}
\begin{split}
p(r_t,x_{t+1} \,|\, x_t,a_t) \approx & \,\mathbb{E}_{z_t \sim p(z_t|x_t)} \,\mathbb{E}_{z_{t+1} \sim p(z_{t+1}|z_t,a_t)}\\
& \bigl[\,p(r_t|z_t,a_t)p(x_{t+1}|z_{t+1})\,\bigr]
\end{split}
\label{eq:latent-expectation}
\end{equation}
where we have made multiple independence assumptions. Especially, we want an
observation encoding model, ${p_\phi(z_t|x_t)}$, that does not depend on the
action (analogous to the ``vision'' component of \citet{world-models}). Second,
we want to predict the next latent variable based on the previous latent
variable and action, i.e., ${p_\phi(z_{t+1}|z_t,a_t)}$, independent of the
observation $x_t$. Furthermore, the reward and next latent variable should not
depend on each other, which allows to predict them using two heads and compute
them in a single neural network pass.

In contrast to \citet{simple}, the policy is conditioned on the latent
variables, so that no decoding into high-dimensional observations is necessary,
\begin{equation}
p(a_t|x_t) \approx
\mathbb{E}_{z_t \sim p(z_t|x_t)}   \bigl[\,p(a_t|z_t)\,\bigr] .
\label{eq:policy-expectation}
\end{equation}

\subsection{Recurrent Dynamics}

Predicting the next latent variable and reward can be improved by introducing a
recurrent variable $h_t$ to the dynamics model, similar to \citet{world-models},
\begin{equation}
\begin{split}
& p(x_{t+1},r_t,h_t|x_t,a_t,h_{t-1})\\
& \approx  \,\mathbb{E}_{z_t \sim p(z_t|x_t)}  \bigl[\, p(h_t|z_t,a_t,h_{t-1}) \,\\
&\phantom{{}\approx{}} \, \mathbb{E}_{z_{t+1} \sim p(z_{t+1}|z_t,a_t,h_t)}  \bigl[\, p(r_t|z_t,a_t,h_t) \\
& \phantom{{}\approx{} \, \mathbb{E}_{z_{t+1} \sim p(z_{t+1}|z_t,a_t,h_t)}  \bigl[\,}p(x_{t+1}|z_{t+1})\,\bigr]\bigr]\\
& = \,\mathbb{E}_{z_t \sim p(z_t|x_t)} \bigl[\, p(h_t|z_t,a_t,h_{t-1})\\
& \phantom{{}={}} \,\mathbb{E}_{z_{t+1} \sim p(z_{t+1}|y_t)}\bigl[\, p(r_t|y_t) \, p(x_{t+1}|z_{t+1})\,\bigr]\bigr].
\label{eq:recurrent-expectation}
\end{split}
\end{equation}
We have made independence assumptions analogous to \cref{eq:latent-expectation}.
In particular, the observation encoder and decoder do not depend on the
recurrent variable, but the latent and reward dynamics do. For notation
purposes, we introduced an intermediate representation $y_t = f(z_t,a_t,h_t)$ in
\cref{eq:recurrent-expectation}, which is a deterministic function of $z_t$,
$a_t$, and $h_t$. The resulting graphical model can be seen in
\cref{fig:graphical-model}.

We can also condition the policy on the recurrent variable, which adds a
dependency from $a_t$ to $h_{t-1}$,
\begin{equation}
p(a_t|x_t,h_{t-1}) \approx \mathbb{E}_{z_t \sim p(z_t|x_t)}
\bigl[ \, p(a_t|z_t,h_{t-1}) \bigr]\!. \label{eq:recurrent-policy}
\end{equation}

From \cref{eq:recurrent-expectation}, \cref{eq:policy-expectation}, and
\cref{eq:recurrent-policy} it becomes clear that several models have to be
learned. We denote the parameters of the world model by $\phi$, and the
parameters of the policy by $\theta$. \cref{tab:models} provides an overview of
all models that need to be learned.

\begin{table}[h]
  \vskip -0.083in
	\caption{Summary of the learned models. Note that $y_t$ is a deterministic
		function of $z_t$, $a_t$, and $h_t$.}
	\label{tab:models}
	\vskip 0.15in
	\begin{center}
		\resizebox{\columnwidth}{!}{
		\begin{tabular}{lll}
			\toprule
			& Observation encoder & $p_\phi(z_t|x_t)$ \\[2pt]
			& Recurrent dynamics & $p_\phi(h_t|z_t,a_t,h_{t-1})$ \\[2pt]
			World model & Reward dynamics & $p_\phi(r_t|y_t)$ \\[2pt]
			& Latent dynamics & $p_\phi(z_{t+1}|y_t)$ \\[2pt]
			& Observation decoder & $p_\phi(\hat{x}_{t+1}|z_{t+1})$ \\
			\midrule
			Agent & Policy & $p_\theta(a_t|z_t)$ or $p_\theta(a_t|z_t,h_{t-1})$ \\
			\bottomrule
		\end{tabular}
    }
	\end{center}
  \vskip -0.15in
\end{table}

\subsection{Architecture}

\paragraph{Latent Representations.}
We use a vector quantized-variational autoencoder \cite{vq-vae} for the
observation encoder and decoder, so each latent variable $z_t$ is a matrix
filled with discrete embedding indices. The observations $x_t$ are the last four
frames of the Atari game, stacked, scaled down to $96 \times 96$ and converted
from RGB to grayscale. Frame stacking allows the observation encoder to
incorporate short-time information, e.g., the velocity of objects, into the
otherwise stationary latent representations.

\begin{figure*}[t]
  \includegraphics[height=5cm]{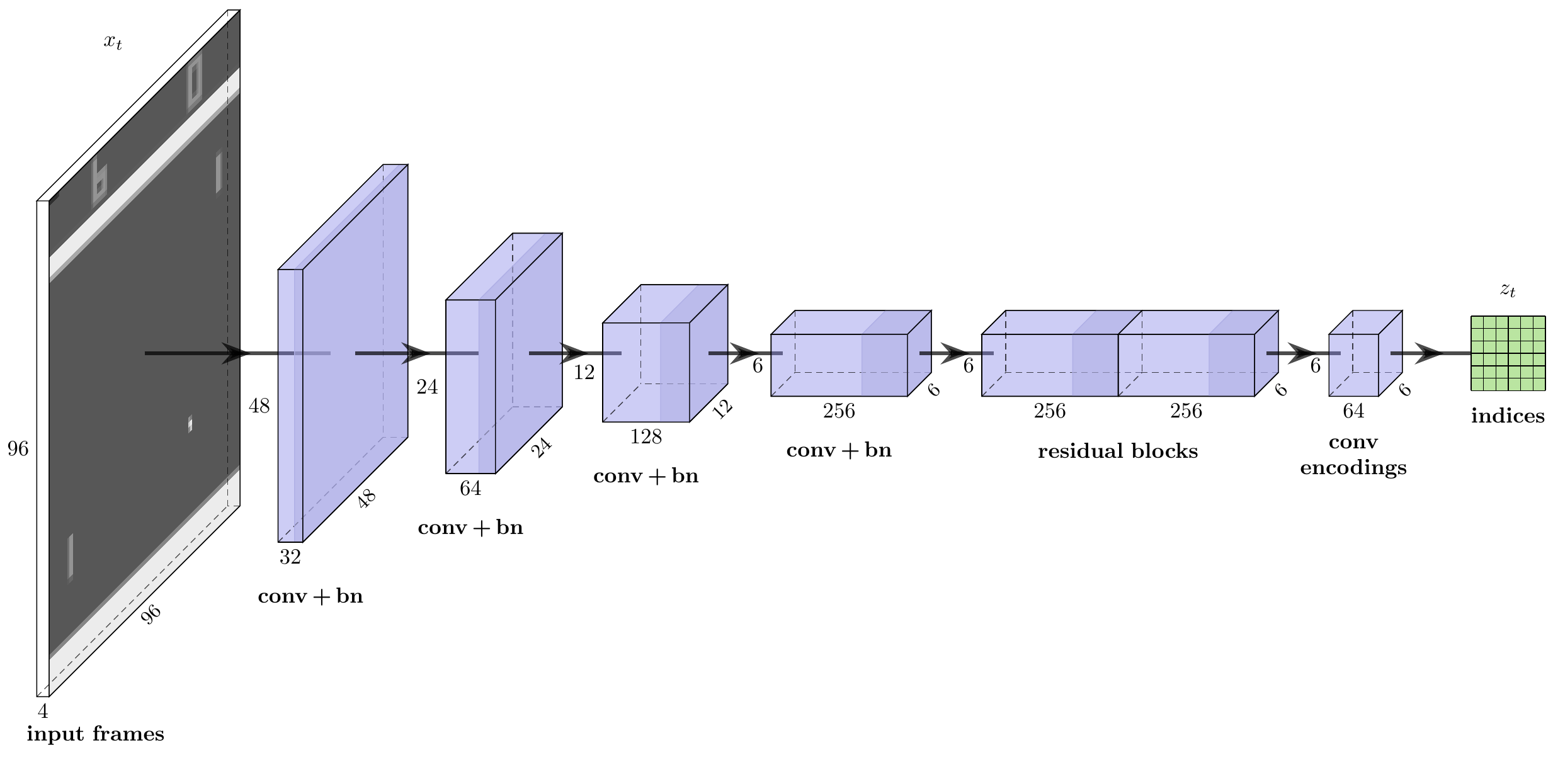}
  \vskip -1.2in
  \hfill
  \includegraphics[height=5cm]{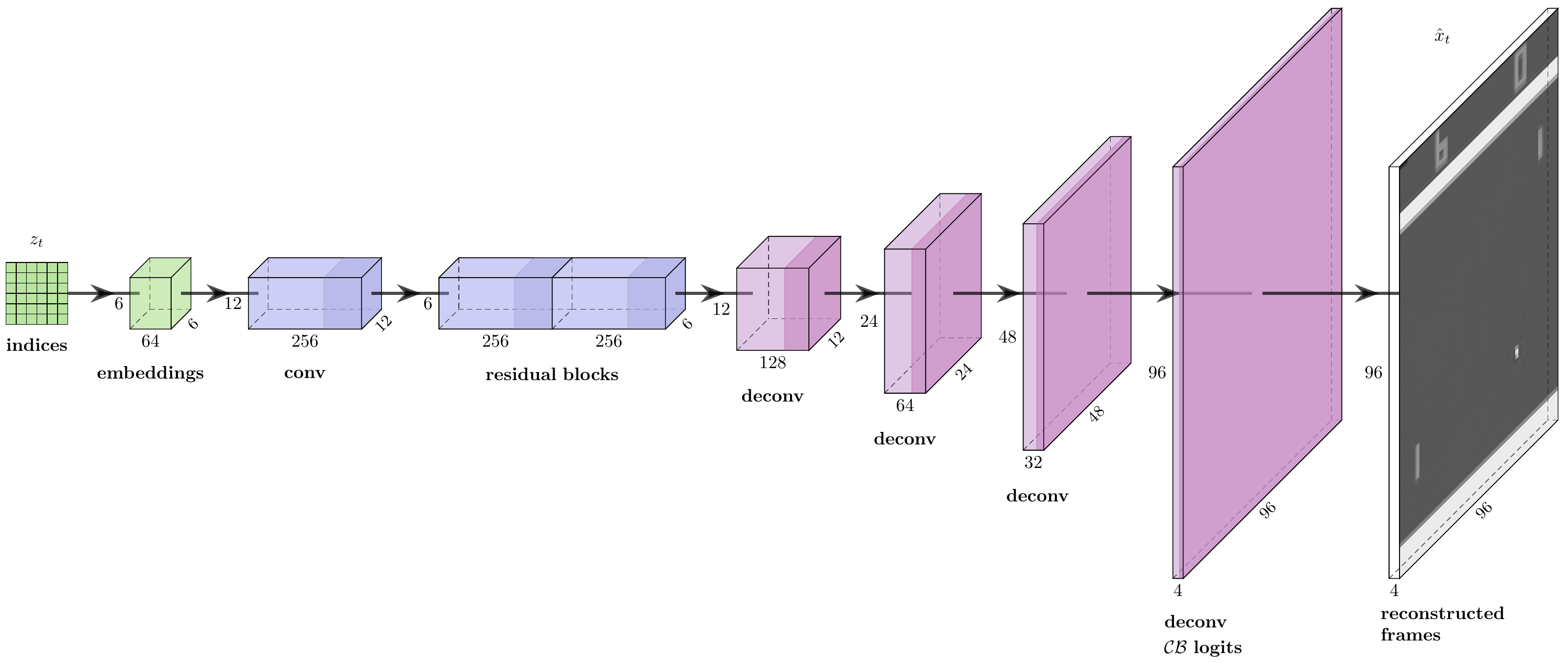}
	\caption{Visualization of the observation encoder architecture (top) and
		decoder architecture (bottom).}
	\label{fig:vqvae}
\end{figure*}

The encoder uses convolutions with batch normalization, while the decoder uses
deconvolutions without batch normalization. All convolutions and deconvolutions
are followed by leaky ReLU nonlinearities (after the batch normalization). We
model the outputs of the decoder with continuous Bernoulli distributions
\cite{continuous-bernoulli} with independence among the stacked frames and
pixels, so the last deconvolution outputs the logits of $96 \times 96 \times 4$
distributions. See \cref{fig:vqvae} for a visualization of the model.

We employ a two-dimensional representation because it can also express local
spatial correlations. Thus, it is better suited to predict the representation at
the next time step, especially when combined with convolutional operations.

\begin{figure*}[t]
	\centering
  \vskip 0.1in
	\includegraphics[width=0.8\linewidth]
	 {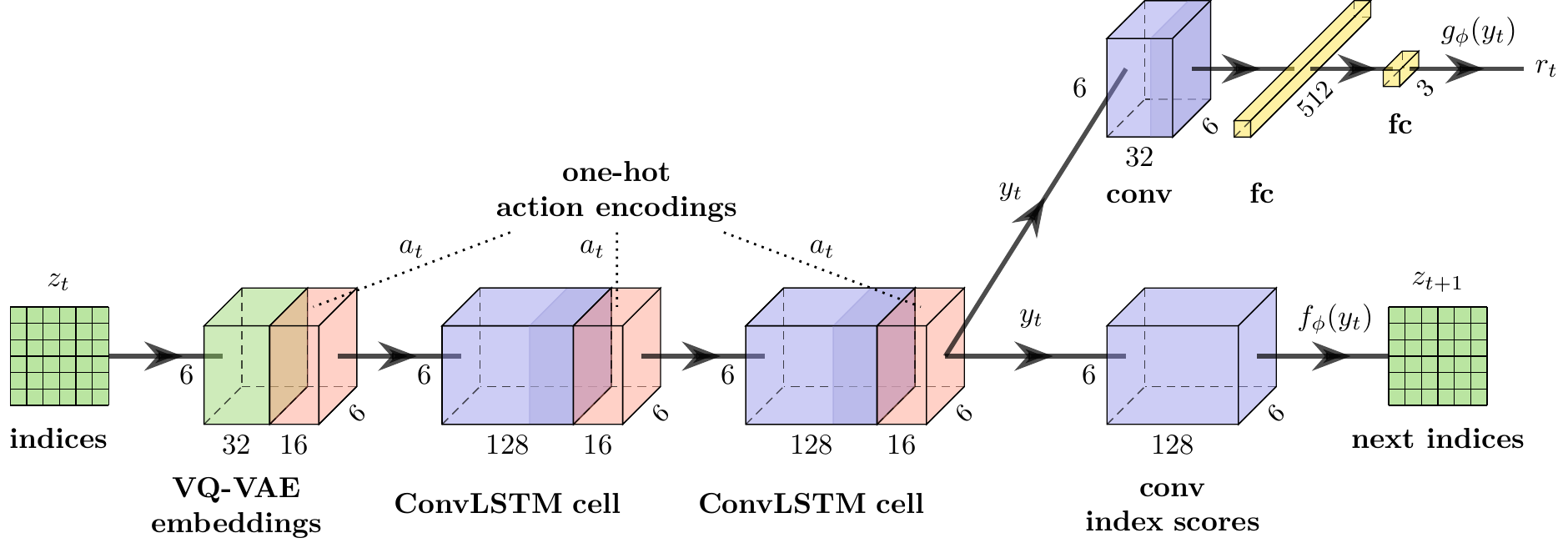}
	\caption{A visualization of the architecture of the dynamics network. After
    the second convolutional LSTM cell the network splits into the reward
    prediction head $g_\phi(y_t)$ at the top and the next latent prediction head
    $f_\phi(y_t)$ at the bottom. The recurrent states $h_t, h_{t-1}$ of the
    LSTM are not visualized for clarity.}
	\label{fig:architecture}
\end{figure*}

\paragraph{Dynamics.}
For the recurrent dynamics we use a two-cell convolutional LSTM
\citep{conv-lstm} with layer normalization. The input consists of a
$6 \times 6 \times 48$ tensor, where the first $32$ channels are the embedding
vectors looked up in the codebook of the VQ-VAE using the indices of the
$6 \times 6$ state representation. The last $16$ channels contain one-hot
encodings of the actions, repeated along the spatial dimensions. By doing this,
we condition the dynamics on the action. The action encodings are also
concatenated to the output of each convolutional LSTM cell, since the action
information might get lost during the forward pass. After the last
convolutional LSTM cell, this corresponds to the intermediate representation
$y_t$ from \cref{fig:graphical-model}, which means that we actually drop its
direct dependence on $z_t$. Then, there are two prediction heads, one for the
next latent variable, $f_\phi(y_t)$ consisting of one convolutional layer, and
one for the reward, $g_\phi(y_t)$ consisting of a convolutional layer and two
fully-connected layers. The convolutional layers are followed by layer
normalization and leaky ReLU nonlinearities. For a detailed depiction of the
model see \cref{fig:architecture}.

The output of $f_\phi(y_t)$ is a $H \times W \times K$ tensor
(${6 \times 6 \times 128}$) that contains the unnormalized scores for the
embedding indices that get normalized via the softmax function,
\begin{equation}
p_\phi\!\left(z_{t+1}^{(j,k)} \,\middle|\, y_t\right)
= \Cat\!\left(K,\, \Softmax\!\left(f_\phi^{(j,k)}(y_t)\right)\right)\!.
\label{eq:latent-dynamics}
\end{equation}
We suppose that the discretization of the latent space stabilizes the dynamics
model, since it has to predict scores for a predefined set of categories instead
of real values, especially considering that the target is moving, i.e., the
latent representations change during training.

The rewards are discretized into three categories $\{-1,0,1\}$ by clipping them
into the interval $[-1,1]$ and rounding them to the nearest integer. The output
of ${g_\phi(y_t)}$ is a $3$-dimensional vector containing the scores for each
reward category, which are also normalized via the softmax function,
\begin{equation}
p_\phi(r_t|y_t) = \Cat\!\left(3, \Softmax\!\left(g_\phi(y_t)\right)\right)\!.
\label{eq:reward-dynamics}
\end{equation}
The support of this distribution is ${r \in \{1,2,3\}}$, so we have to map the
rewards accordingly (${r = r_\text{orig} + 2}$) when we compute the likelihood.

\paragraph{Policy.}
The input of the policy network are the embedding vectors and it processes them
using two convolutional layers with layer normalization, followed by a fully
connected layer. Unlike \cref{eq:recurrent-policy} the policy does not depend on
the recurrent variable $h_{t-1}$ from the world model in our experiments, but
this could be useful for more complex environments. The output of the network
${f_\theta(z_t)}$ is again a vector of unnormalized scores of a categorical
distribution for the $M$ possible actions,
\begin{equation}
p_\theta(a_t|z_t)
= \Cat\!\left(M, \Softmax\!\left(f_\theta(z_t)\right)\right)\!.
\end{equation}

\begin{table}[t]
	\centering
  \vskip -0.083in
	\caption{Number of parameters of the world model compared with \citet{simple}
    (their number is approximate).}
	\label{tab:parameters}
	\vskip 0.15in
  \resizebox{0.42\columnwidth}{!}{
	\begin{tabular}{lr}
		\toprule
		Model & \# parameters \\
		\midrule
		Ours & \num{10332740} \\
		SimPLe & \num{74000000} \\
		\bottomrule
	\end{tabular}
  }
  \vskip -0.1in
\end{table}

\paragraph{Number of Parameters.} \cref{tab:parameters} shows the number of
parameters of our model compared with the model by \citet{simple}, which
uses about seven times as many parameters. \cref{tab:parameters-detail} shows
the number of parameters of our models in detail. At training time all models
are used, but at test time only the encoder and the policy network are required.

\subsection{Training}

The distributions in \cref{eq:latent-dynamics} and \cref{eq:reward-dynamics} are
trained using maximum likelihood. The policy is optimized on simulated
experience using proximal policy optimization \cite{ppo}, a model-free
reinforcement learning algorithm. We approximate the expectations in
\cref{eq:latent-expectation} and \cref{eq:policy-expectation} with single Monte
Carlo samples. While we simulate experience, we use the same sample
${z_t \sim p(z_t|x_t)}$ for both the dynamics, \cref{eq:recurrent-expectation},
and the policy, \cref{eq:policy-expectation}. At inference time, when the policy
is applied to a real environment, \cref{eq:policy-expectation} still needs
access to the observation encoder ${p_\phi(z_t|x_t)}$ in order to sample the
latent variables.

\begin{table}[t]
	\centering
  \vskip -0.084in
	\caption{Number of parameters of our models in detail. The encoder and decoder
  both need access to the embedding vectors, therefore the sum of their
  individual number of parameters is slightly higher than for the total VQ-VAE.}
  \label{tab:parameters-detail}
	\vskip 0.15in
  \resizebox{0.81\columnwidth}{!}{
	\begin{tabular}{lr}
		\toprule
		Model & \# parameters \\
		\midrule
		World model & \num{10332740} \\
		VQ-VAE & \num{4089313} \\
		Encoder & \num{2014720} \\
		Decoder & \num{2078689} \\
		Dynamics network & \num{6243427} \\
		Policy network & \num{1297415} \\
		\midrule
		World model + policy (training) & \num{11630155} \\
		Encoder + policy (inference) & \num{3312135} \\
		\bottomrule
	\end{tabular}
  }
\vskip -0.1in
\end{table}

\paragraph{Episodic Environments.}
Atari games are episodic, therefore the world model needs to predict terminal
states, for instance by predicting a binary variable that indicates the end of
the episode. This prediction has to be reliable, since an incorrect prediction
of ``true'' can have a severe impact on the simulated experience and thus on
the policy.

We follow \citet{simple} and end all episodes after a fixed number of steps
(e.g., 50), so the world model does not have to terminate episodes.

Furthermore, we adopt the idea of randomly selecting the initial observation of
a simulated episode from the collected real data. This enables the policy to
learn from experience from any stage of the environment, although the number of
time steps is limited. On the downside, this prevents the policy to learn from
effects that are longer than the fixed number of steps \cite{simple}.

\paragraph{Iterative Training.}
We adopt the iterative training procedure from \citet{simple} and alternate
between interacting with the real environment, training the world model, and
training the policy.

We also adopt the number of interactions per iteration,
\num{6400}, and the number of iterations, \num{15}. The authors state that they
perform additional \num{6400} interactions prior to the first iteration. Thus,
we perform \num{12800} interactions in the first iteration, resulting in the
same number of total interactions, ${\num{12800} + \num{6400} \times \num{14} = \num{102400}}$.
In the first iteration we use a random uniform policy.

\paragraph{Warming Up Latent Representations.}
After collecting the first batch of data, we train the VQ-VAE separately for
\num{50} epochs with a higher learning rate. We want to give the dynamics model
a better starting point, with representations that already contain useful
information. This cannot be done in later training stages, as the dynamics
model would not be able to keep up with the representations.

\paragraph{Fixed Representations.}
After warming up, we even slow down the state representation training by
updating the parameters of the VQ-VAE only in every second training step, so the
targets of the dynamics model are moving slower.

\paragraph{Reward Loss.}
If the gradients coming from the reward prediction head have a high magnitude,
they have a degrading effect on the performance of the next latent prediction
head. We solve this issue by scaling down the cross-entropy loss of the rewards
to reduce its influence on the entire model, but using a higher learning rate
for the reward prediction head to compensate for the smaller gradients.

\paragraph{Constant KL Term.}
\citet{simple} state that the weight of the KL divergence loss of a variational
autoencoder is game dependent, which makes VAEs impractical to apply on all
Atari games without fine-tuning. The VQ-VAE does not suffer from this problem,
since the KL term is constant and depends on the number of embeddings.

\begin{figure*}
  \vskip 0.02in
	\begin{subfigure}[t]{0.33\linewidth}
		\centering
		\includegraphics[width=\columnwidth,trim={0 0 0 0.5cm},clip]
		{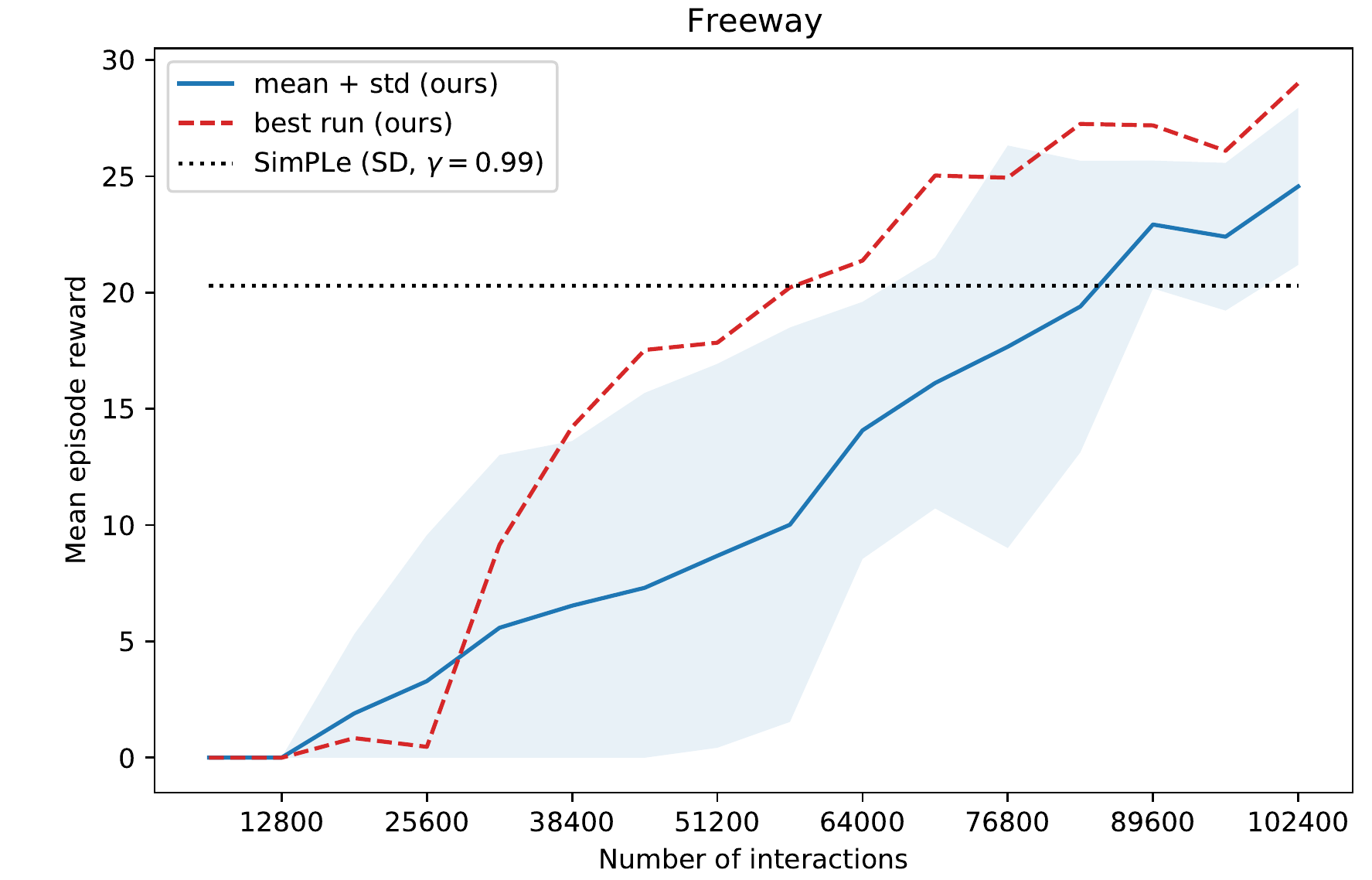}
		\caption{Freeway}
		\label{fig:training-plot-freeway}
	\end{subfigure}
	\begin{subfigure}[t]{0.33\linewidth}
		\centering
		\includegraphics[width=\columnwidth,trim={0 0 0 0.5cm},clip]
		{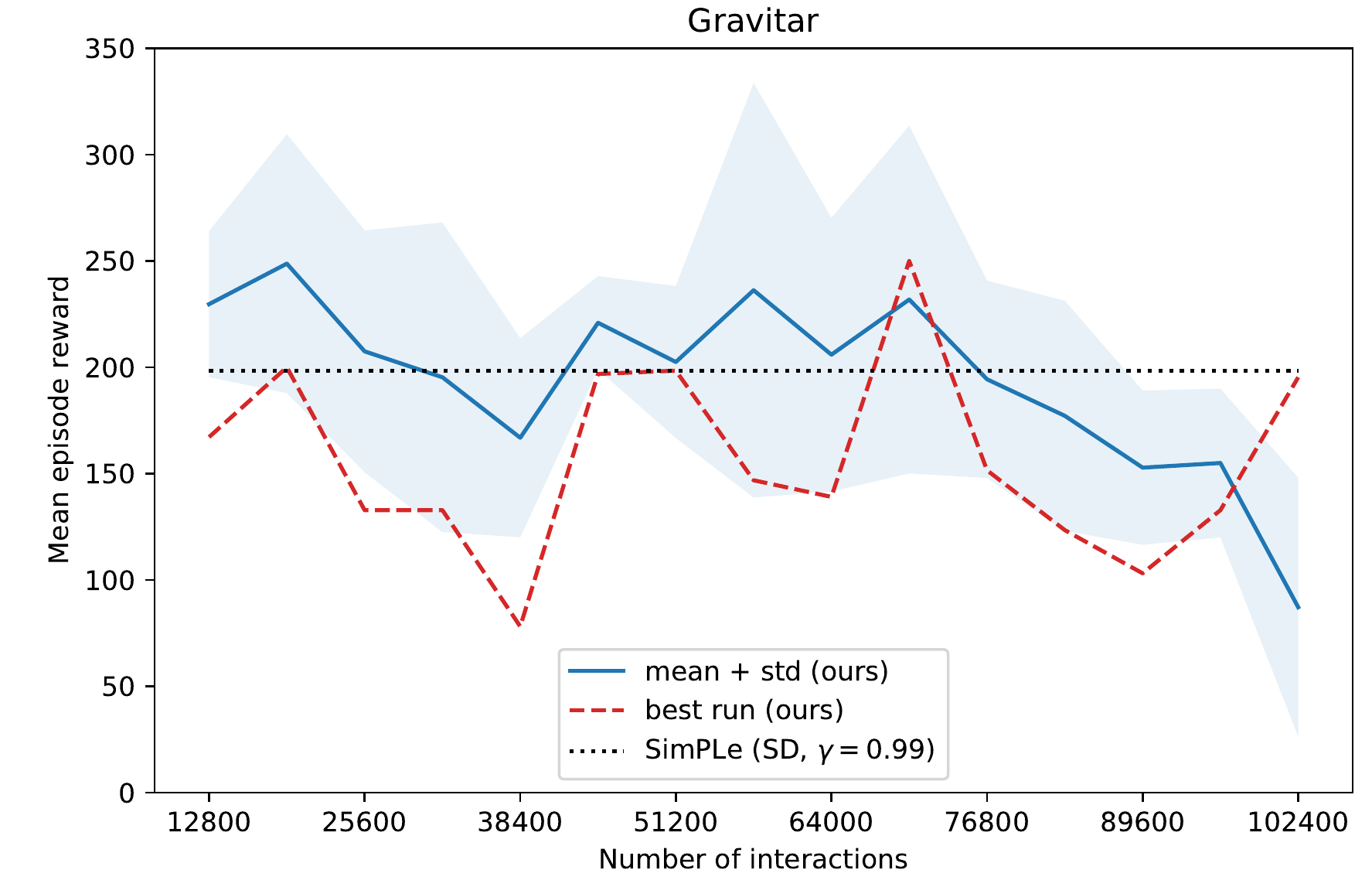}
		\caption{Gravitar}
		\label{fig:training-plot-gravitar}
	\end{subfigure}
	\begin{subfigure}[t]{0.33\linewidth}
		\centering
		\includegraphics[width=\columnwidth,trim={0 0 0 0.5cm},clip]
		{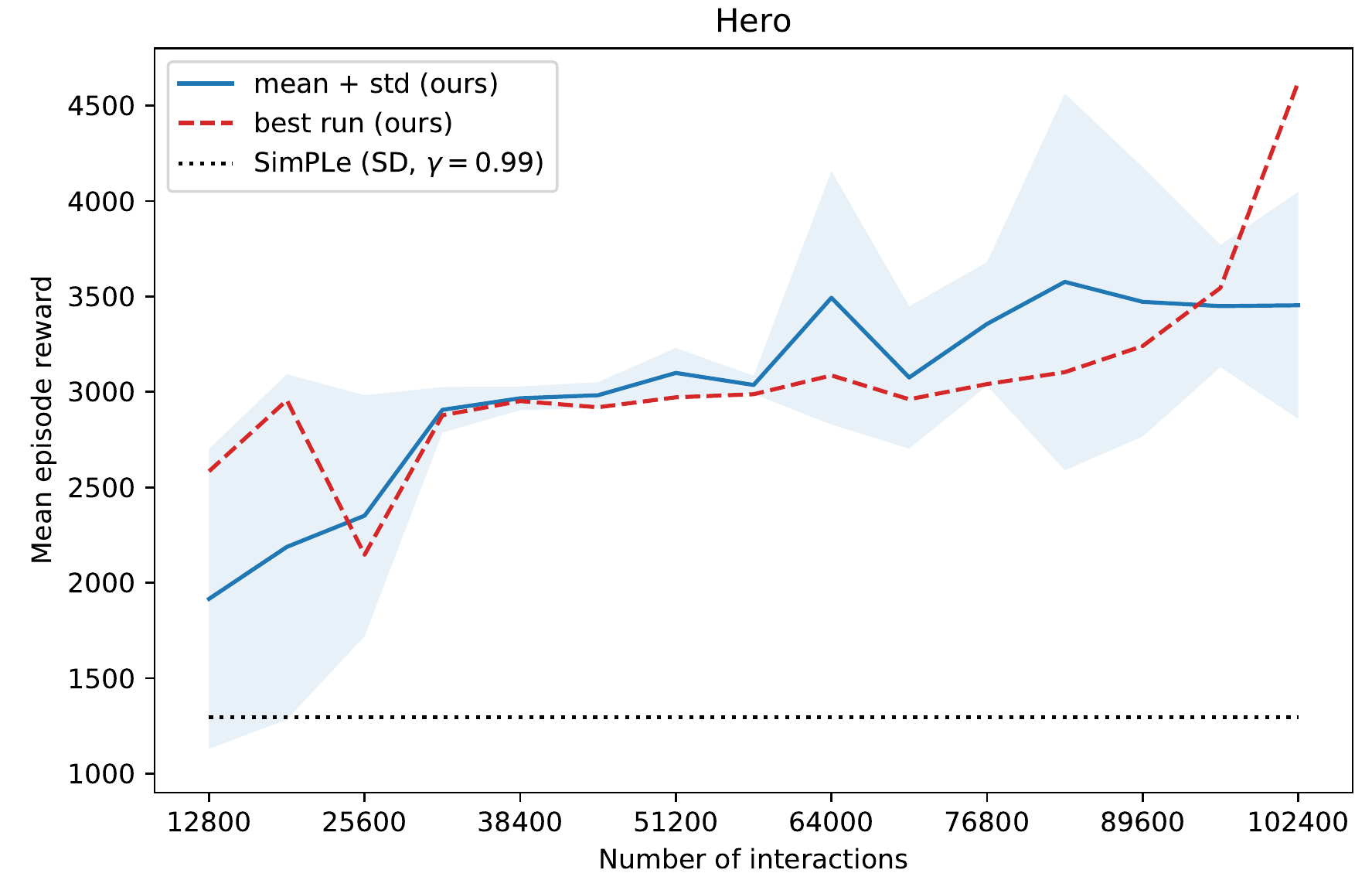}
		\caption{Hero}
		\label{fig:training-plot-hero}
	\end{subfigure}
	\caption{Mean episode reward across five training runs for three Atari
		environments. The x-axis shows the number of interactions with the real
		environment, and does not reflect the number of parameter updates that were
		performed in between. For SimPLe \citep{simple} we only know the final score
		which is depicted by a straight line.}
	\label{fig:training-plots}
\end{figure*}

\section{Evaluation}

\begin{figure}[t]
  \centering
  \vspace{0.055in}
  \begin{subfigure}[t]{0.48\columnwidth}
    \centering
    \includegraphics[height=3.5cm]
    {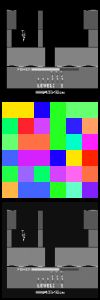}
    \includegraphics[height=3.5cm]
    {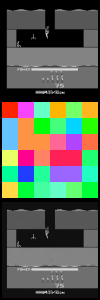}
    \includegraphics[height=3.5cm]
    {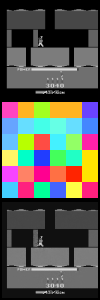}
    \caption{Hero}
    \label{fig:latent-frame-hero}
  \end{subfigure}
  \hfill
  \begin{subfigure}[t]{0.48\columnwidth}
    \centering
    \includegraphics[height=3.5cm]
    {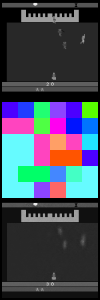}
    \includegraphics[height=3.5cm]
    {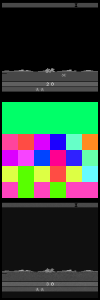}
    \includegraphics[height=3.5cm]
    {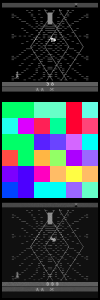}
    \caption{Krull}
    \label{fig:latent-frame-krull}
  \end{subfigure}
  \caption{Visualization of original game frames (top row), the encoded
 	$6 \times 6$ discrete latent variables (middle row), and the reconstructions
 	from the VQ-VAE (bottom row). We assign colors to the embedding indices
 	and draw a colored square for each entry of the latent matrix. A square does
  not necessarily correspond to the same area in the frame as they have larger
  receptive fields.}
 \label{fig:latent-frames}
\end{figure}

We compare our method with the variant of \citet{simple} (stochastic discrete,
50 steps, $\gamma = 0.99$) that comes closest to our model in terms of
hyperparameters (discount rate, batch size etc.) and number of parameter
updates.

This implies that the results of \citet{simple} that are shown are not
necessarily the best across all of their variants, but the best for a fairer
comparison with our method. We do not know the performance difference to
\citet{simple} in terms of training time and test time, but considering the
model size our method should be a lot faster.

We restrict our agent to $100K$ interactions with the environment and average
the results over five training runs. For every run we evaluate the latest policy
in each iteration by rolling out $32$ episodes in the real environment and
computing the mean of the (cumulative) episode rewards. In
\cref{tab:results-mean} we report the mean final episode rewards (i.e., the mean
episode reward after the final iteration; averaged over five runs) for $36$
Atari environments. Our method achieves a higher value than SimPLe \cite{simple}
in $20$ out of $36$ environments (when also considering no frame stacking, as
explained below).

\paragraph{Learning Curves.}
\cref{fig:training-plots} shows three cases of learning curves that were typical
for our model. First, \cref{fig:training-plot-freeway} shows an example of an
environment in which the agent's performance successfully increases over the
course of training. Secondly, \cref{fig:training-plot-gravitar} shows an
example of an environment in which the agent's performance \textit{decreases}
over the course of training. This can have various reasons, e.g., when the
agent reaches a new area of the state space and the environment dynamics change
drastically. Finally, \cref{fig:training-plot-hero} shows an example of an
environment in which the agent has comparably high performance starting from
the first iteration, which is most likely due to model bias.

\paragraph{Latent Representations.}
In \cref{fig:latent-frames} we can see that the state representation model is
able to encode almost all information into the latent representation. However,
the embedding indices are tuned for the decoder, so the dynamics model still has
a hard task. We picked these two environments to show that changes in the scene
(\cref{fig:latent-frame-hero}) or even switching scenes
(\cref{fig:latent-frame-krull}) can be represented, although this can cause some
loss of details (e.g., see the left reconstruction in
\cref{fig:latent-frame-krull}).

\paragraph{No Frame Stacking.} Our default model stacks the last four
frames to incorporate short-time information into the state
representations. On the downside, this introduces complexity, since the
same frame can get a different representation, depending on the three
other frames in the stack. This in turn will make it harder for the
dynamics model to predict the next representation. The results show that no
frame stacking improves the performance in some environments, as can be seen in
\cref{tab:results-mean}.

\paragraph{Exploration.}
In our experiments we observe the same phenomenon as \citet{simple}, namely that
the results can vary drastically for different runs with the same
hyperparameters but different random seeds. The main reasons might be that
the world model cannot infer dynamics of regions of the environment's state
space that it has never seen before, and that the algorithm is very sensitive to
the exploration-exploitation trade-off, as the number of interactions is low.

\clearpage

\begin{table*}[t]
	\centering
  \vspace{-0.1in}
	\caption{Comparison of our method (with and without frame stacking) with
    SimPLe \citep{simple} and model-free PPO \citep{ppo} trained with $100K$
    steps. Our scores are the mean final episode reward, averaged over five runs
    $\pm$ standard deviation. The PPO scores are taken from \citet{simple}.}
  \label{tab:results-mean}
  \vskip 0.1in
  \resizebox{\linewidth}{!}{
  \begin{tabular}{lrlrlrlrl}
    \toprule
    & \multicolumn{2}{c}{Ours (4 frames)}
    & \multicolumn{2}{c}{Ours (1 frame)}
    & \multicolumn{2}{c}{SimPLe (SD, $\gamma = 0.99$)}
    & \multicolumn{2}{c}{PPO $100K$} \\
    \midrule
    Game & Mean & Std. Dev. & Mean & Std. Dev. & Mean & Std. Dev.  & Mean & Std. Dev. \\
    \midrule
    Alien
    & $409.9$ & $(\pm \, 73.0)$
    & $\mathbf{423.3}$ & $(\pm \, 48.1)$
    & $405.2$ & $(\pm \, 130.8)$
    & $291.0$ & $(\pm \, 40.3)$ \\
    Amidar
    & $37.6$ & $(\pm \, 13.6)$
    & $30.5$ & $(\pm \, 10.1)$
    & $\mathbf{88.0}$ & $(\pm \, 23.8)$
    & $56.5$ & $(\pm \, 20.8)$ \\
    Assault
    & $375.4$ & $(\pm \, 111.9)$
    & $408.3$ & $(\pm \, 27.8)$
    & $369.3$ & $(\pm \, 107.8)$
    & $\mathbf{424.2}$ & $(\pm \, 55.8)$ \\
    Asterix
    & $504.4$ & $(\pm \, 53.3)$
    & $456.5$ & $(\pm \, 146.4)$
    & $\mathbf{1089.5}$ & $(\pm \, 335.3)$
    & $385.0$ & $(\pm \, 104.4)$ \\
    Asteroids
    & $862.9$ & $(\pm \, 85.4)$
    & $989.9$ & $(\pm \, 88.7)$
    & $731.0$ & $(\pm \, 165.3)$
    & $\mathbf{1134.0}$ & $(\pm \, 326.9)$ \\
    Atlantis
    & $9413.1$ & $(\pm \, 3349.8)$
    & $15463.7$ & $(\pm \, 5478.7)$
    & $14481.6$ & $(\pm \, 2436.9)$
    & $\mathbf{34316.7}$ & $(\pm \, 5703.8)$ \\
    BankHeist
    & $101.2$ & $(\pm \, 17.4)$
    & $\mathbf{249.3}$ & $(\pm \, 49.8)$
    & $8.2$ & $(\pm \, 4.4)$
    & $16.0$ & $(\pm \, 12.4)$ \\
    BattleZone
    & $\mathbf{5631.2}$ & $(\pm \, 1179.1)$
    & $5531.3$ & $(\pm \, 2515.4)$
    & $5184.4$ & $(\pm \, 1347.5)$
    & $5300.0$ & $(\pm \, 3655.1)$ \\
    BeamRider
    & $410.4$ & $(\pm \, 55.4)$
    & $527.6$ & $(\pm \, 61.8)$
    & $422.7$ & $(\pm \, 103.6)$
    & $\mathbf{563.6}$ & $(\pm \, 189.4)$ \\
    Bowling
    & $27.9$ & $(\pm \, 4.8)$
    & $24.5$ & $(\pm \, 5.1)$
    & $\mathbf{34.4}$ & $(\pm \, 16.3)$
    & $17.7$ & $11.2$ \\
    Boxing
    & $-2.8$ & $(\pm \, 5.7)$
    & $-9.3$ & $(\pm \, 12.6)$
    & $\mathbf{9.1}$ & $(\pm \, 8.8)$
    & $-3.9$ & $(\pm \, 6.4)$ \\
    Breakout
    & $8.8$ & $(\pm \, 1.5)$
    & $8.4$ & $(\pm \, 1.5)$
    & $\mathbf{12.7}$ & $(\pm \, 3.8)$
    & $5.9$ & $(\pm \, 3.3)$ \\
    ChopperCommand
    & $766.2$ & $(\pm \, 195.3)$
    & $590.6$ & $(\pm \, 335.0)$
    & $\mathbf{1246.9}$ & $(\pm \, 392.0)$
    & $730.0$ & $(\pm \, 199.0)$ \\
    CrazyClimber
    & $\mathbf{47536.9}$ & $(\pm \, 6114.9)$
    & $36923.8$ & $(\pm \, 2780.6)$
    & $39827.8$ & $(\pm \, 22582.6)$
    & $18400.0$ & $(\pm \, 5275.1)$ \\
    DemonAttack
    & $195.0$ & $(\pm \, 76.4)$
    & $\mathbf{211.3}$ & $(\pm \, 86.2)$
    & $169.5$ & $(\pm \, 41.8)$
    & $192.5$ & $(\pm \, 83.1)$ \\
    FishingDerby
    & $-89.6$ & $(\pm \, 4.5)$
    & $\mathbf{-87.9}$ & $(\pm \, 4.1)$
    & $-91.5$ & $(\pm \, 2.8)$
    & $-95.6$ & $(\pm \, 4.3)$ \\
    Freeway
    & $\mathbf{24.6}$ & $(\pm \, 3.4)$
    & $11.3$ & $(\pm \, 9.6)$
    & $20.3$ & $(\pm \, 18.5)$
    & $8.0$ & $(\pm \, 9.8)$ \\
    Frostbite
    & $214.4$ & $(\pm \, 10.2)$
    & $219.1$ & $(\pm \, 45.6)$
    & $\mathbf{254.7}$ & $(\pm \, 4.9)$
    & $174.0$ & $(\pm \, 40.7)$ \\
    Gopher
    & $687.2$ & $(\pm \, 91.1)$
    & $\mathbf{1398.4}$ & $(\pm \, 166.5)$
    & $771.0$ & $(\pm \, 160.2)$
    & $246.0$ & $(\pm \, 103.3)$ \\
    Gravitar
    & $87.2$ & $(\pm \, 60.9)$
    & $82.2$ & $(\pm \, 64.5)$
    & $198.3$ & $(\pm \, 39.9)$
    & $\mathbf{235.0}$ & $(\pm \, 197.2)$ \\
    Hero
    & $3453.6$ & $(\pm \, 594.7)$
    & $\mathbf{3911.6}$ & $(\pm \, 1259.3)$
    & $1295.1$ & $(\pm \, 1600.1)$
    & $569.0$ & $(\pm \, 1100.9)$ \\
    IceHockey
    & $-13.6$ & $(\pm \, 2.5)$
    & $-12.0$ & $(\pm \, 2.1)$
    & $-10.5$ & $(\pm \, 2.2)$
    & $\mathbf{-10.0}$ & $(\pm \, 2.1)$ \\
    Jamesbond
    & $66.6$ & $(\pm \, 7.8)$
    & $46.6$ & $(\pm \, 24.2)$
    & $\mathbf{125.3}$ & $(\pm \, 112.5)$
    & $65.0$ & $(\pm \, 46.4)$ \\
    Kangaroo
    & $245.0$ & $(\pm \, 99.6)$
    & $276.3$ & $(\pm \, 136.7)$
    & $\mathbf{323.1}$ & $(\pm \, 359.8)$
    & $140.0$ & $(\pm \, 102.0)$ \\
    Krull
    & $3520.2$ & $(\pm \, 211.4)$
    & $3241.0$ & $(\pm \, 448.4)$
    & $\mathbf{4539.9}$ & $(\pm \, 2470.4)$
    & $3750.4$ & $(\pm \, 3071.9)$ \\
    KungFuMaster
    & $11903.1$ & $(\pm \, 4399.5)$
    & $8521.2$ & $(\pm \, 1330.9)$
    & $\mathbf{17257.2}$ & $(\pm \, 5502.6)$
    & $4820.0$ & $(\pm \, 983.2)$ \\
    MsPacman
    & $652.2$ & $(\pm \, 92.6)$
    & $668.3$ & $(\pm \, 86.4)$
    & $\mathbf{762.8}$ & $(\pm \, 331.5)$
    & $496.0$ & $(\pm \, 379.8)$ \\
    NameThisGame
    & $\mathbf{2448.4}$ & $(\pm \, 179.5)$
    & $2119.4$ & $(\pm \, 217.9)$
    & $1990.4$ & $(\pm \, 284.7)$
    & $2225.0$ & $(\pm \, 423.7)$ \\
    Pong
    & $\mathbf{11.8}$ & $(\pm \, 6.9)$
    & $-3.9$ & $(\pm \, 7.6)$
    & $5.2$ & $(\pm \, 9.7)$
    & $-20.5$ & $(\pm \, 0.6)$ \\
    PrivateEye
    & $\mathbf{99.4}$ & $(\pm \, 1.2)$
    & $96.9$ & $(\pm \, 5.4)$
    & $58.3$ & $(\pm \, 45.4)$
    & $10.0$ & $(\pm \, 20.0)$ \\
    Qbert
    & $480.9$ & $(\pm \, 143.5)$
    & $\mathbf{617.5}$ & $(\pm \, 149.5)$
    & $559.8$ & $(\pm \, 183.8)$
    & $362.5$ & $(\pm \, 117.8)$ \\
    Riverraid
    & $2100.2$ & $(\pm \, 50.4)$
    & $\mathbf{2273.3}$ & $(\pm \, 188.5)$
    & $1587.0$ & $(\pm \, 818.0)$
    & $1398.0$ & $(\pm \, 513.8)$ \\
    RoadRunner
    & $1562.5$ & $(\pm \, 440.2)$
    & $1723.8$ & $(\pm \, 688.2)$
    & $\mathbf{5169.4}$ & $(\pm \, 3939.0)$
    & $1430.0$ & $(\pm \, 760.0)$ \\
    Seaquest
    & $458.1$ & $(\pm \, 155.7)$
    & $\mathbf{531.5}$ & $(\pm \, 105.1)$
    & $370.9$ & $(\pm \, 128.2)$
    & $370.0$ & $(\pm \, 103.3)$ \\
    UpNDown
    & $1128.2$ & $(\pm \, 247.6)$
    & $1354.6$ & $(\pm \, 741.4)$
    & $2152.6$ & $(\pm \, 1192.4)$
    & $\mathbf{2874.0}$ & $(\pm \, 1105.8)$ \\
    YarsRevenge
    & $4096.0$ & $(\pm \, 520.9)$
    & $4360.3$ & $(\pm \, 1156.9)$
    & $2980.2$ & $(\pm \, 778.6)$
    & $\mathbf{5182.0}$ & $(\pm \, 1209.3)$ \\
    \bottomrule
  \end{tabular}
  }
  \vskip -0.1in
\end{table*}

\clearpage

\paragraph{LSTM architecture.}
Instead of a convolutional LSTM, we also tried follow-up architectures like the
spatio-temporal LSTM \citep{st-lstm} or causal LSTM \citep{causal-lstm}, but
for our problem the performance gain was not significant enough, compared with
the imposed additional training time and number of parameters.

\section{Limitations and Future Work}

Currently, our dynamics model samples the indices in the latent representation
independently. This might be disadvantageous because conditional dependencies
between the indices, which correspond to certain areas of the video frames, are
ignored. So in the future it would be interesting to predict them
autoregressively, e.g., with a conditional PixelCNN \citep{conditional-pixelcnn}
conditioned on $y_t$, to see whether this solves prediction errors like
duplication or incoherent movement of objects. Nevertheless, an autoregressive
model might have a negative impact on the training times (when training the
dynamics model and when simulating experience), and it has to be seen whether
the resulting training times are acceptable.

Another line of research should improve exploration in order to enable even
higher sample efficiency. Furthermore, we would like the world model to predict
terminal states and move away from terminating episodes after a fixed number of
steps, since this introduces a new hyperparameter that needs to be tuned, and
prevents the agent from learning beyond this time horizon.

\section{Conclusion}

In this paper we demonstrate that a generative model with a discrete latent
space can be used to strongly decrease the size of world models. We employ a
VQ-VAE with a discrete two-dimensional latent space. We show that this powerful
model is able to effectively encode the complex visual observations of Atari
games. The chosen model and latent space have produced representations that are
more stable and expressive at the same time. Our experiments show that acting
entirely in latent space is possible, which speeds up training since no decoding
into high-dimensional frames is required.

\bibliography{bibliography}
\bibliographystyle{icml2021}


\end{document}